\documentclass[10pt,twocolumn,letterpaper]{article}

\usepackage{iccv,times,graphicx,tabulary,multirow,subfig,xspace,xcolor}
\usepackage{amsthm,amssymb,fixmath,mathtools,nicefrac}
\usepackage[pagebackref=true,breaklinks=true,letterpaper=true,colorlinks,
  citecolor=citecolor,bookmarks=false]{hyperref}
\definecolor{citecolor}{RGB}{34,139,34}

\usepackage{booktabs} 
\usepackage{multirow}
\usepackage{array}

\begin{document}
\title{Human Keypoint Detection by Progressive Context Refinement}
\track{COCO Keypoint Detection}

\author{Jing Zhang, Zhe Chen, and Dacheng Tao\\
UBTECH Sydney AI Centre, School of Computer Science, Faculty of Engineering\\
The University of Sydney, Darlington, NSW 2008, Australia\\
{\tt\small \{jing.zhang1, zhe.chen1, dacheng.tao\}@sydney.edu.au}
}

\maketitle

\begin{abstract}
Human keypoint detection from a single image is very challenging due to occlusion, blur, illumination and scale variance of person instances. In this paper, we find that context information plays an important role in addressing these issues, and propose a novel method named progressive context refinement (PCR) for human keypoint detection. First, we devise a simple but effective context-aware module (CAM) that can efficiently integrate spatial and channel context information to aid feature learning for locating hard keypoints. Then, we construct the PCR model by stacking several CAMs sequentially with shortcuts and employ multi-task learning to progressively refine the context information and predictions. Besides, to maximize PCR's potential for the aforementioned hard case inference, we propose a hard-negative person detection mining strategy together with a joint-training strategy by exploiting the unlabeled coco dataset and external dataset. Extensive experiments on the COCO keypoint detection benchmark demonstrate the superiority of PCR over representative state-of-the-art (SOTA) methods. Our single model achieves comparable performance with the winner of the 2018 COCO Keypoint Detection Challenge. The final ensemble model sets a new SOTA on this benchmark.
\end{abstract}

\section{Introduction}
\label{sec:intro}
\begin{figure}[t]
\centering
\includegraphics[width=1\linewidth]{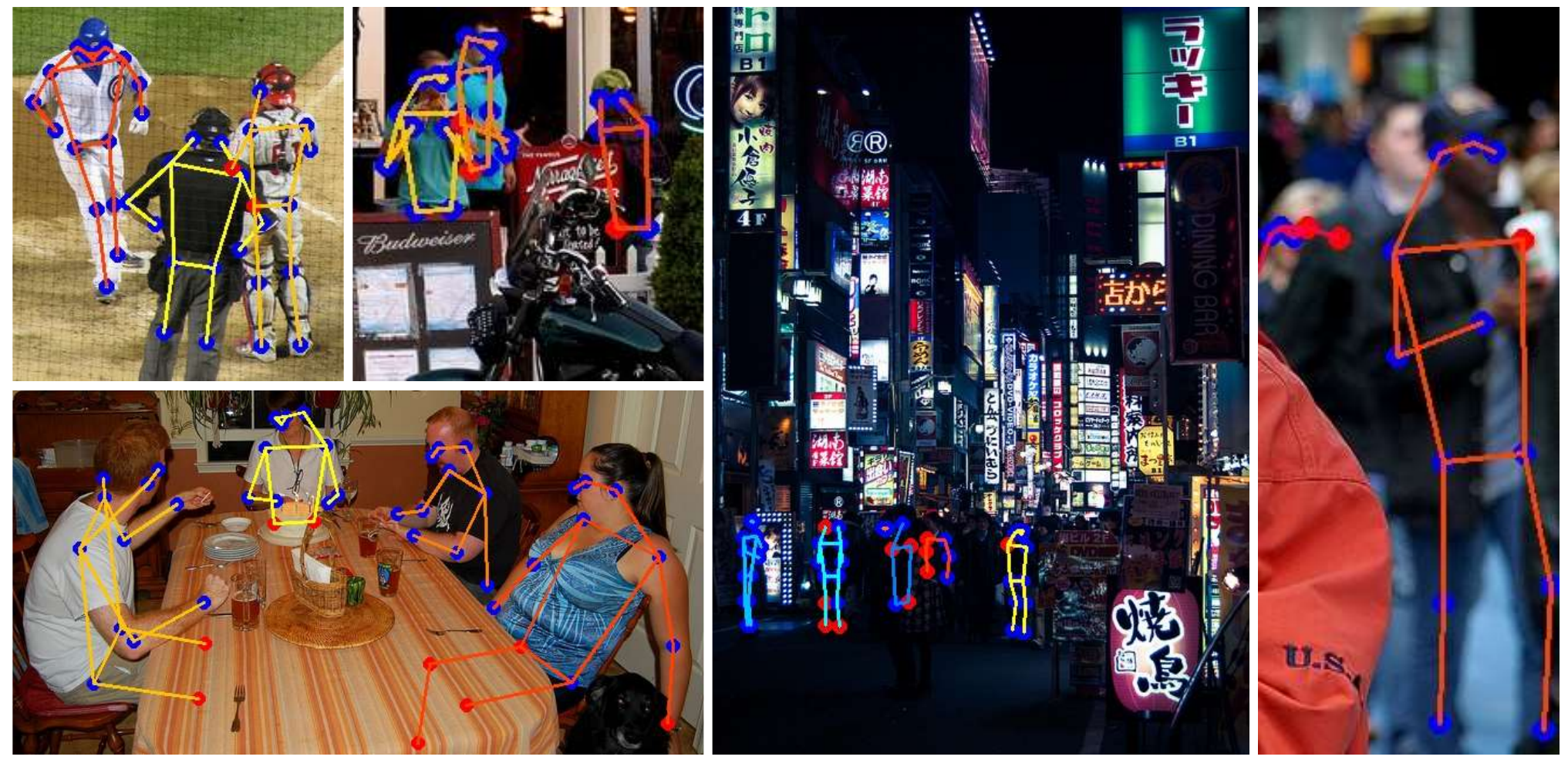}
\caption{Some examples from the MS COCO dataset \cite{lin2014microsoft}, where occluded, under-exposed and blurry person instances are very common. Blue and red dots denote the annotated visible and invisible keypoints, respectively.}
\label{fig:coco_observation}
\end{figure}

Human keypoint detection is also known as human pose estimation (HPE) refers to detecting keypoints' location and recognizing their categories for each person instance from a given image. It is very useful in many downstream applications such as activity recognition, human-robot interaction, and video surveillance. However, HPE is very challenging even for human annotators. For example, 35\% keypoints are unannotated in the COCO training dataset \cite{lin2014microsoft} due to various factors including occlusion, truncation, under-exposed imaging, blurry appearance and low-resolution of person instances. Some examples are shown in Figure~\ref{fig:coco_observation}.

Prior methods have made significant progress in this area with the success of deep convolutional neural networks (DCNNs) \cite{toshev2014deeppose,newell2016stacked,xiao2018simple,chen2018cascaded,sun2019deep}. Toshev and Szegedy propose one of the pioneer DCNNs-based work named DeepPose for HPE \cite{toshev2014deeppose}, which directly learns body part coordinates from an image. Instead, Heatmap based representation has gained prominence in follow-up studies, which represents the keypoint location by placing a 2D Gaussian probability density map at each corresponding coordinate. Newell \etal proposed the well-known hourglass module to learn the heatmaps \cite{newell2016stacked}, which is a fully convolutional architecture. Chen \etal proposed Cascaded Pyramid Network (CPN) to learn a feature pyramid in the first component GlobalNet and handle difficult keypoints by the second component RefineNet \cite{chen2018cascaded}. Recently, Xiao \etal proposed a simple baseline for HPE by using a simple deconvolutional decoder \cite{xiao2018simple}. Sun \etal propose the High-resolution Net (HRNet) which aims for learning deep high-resolution feature representation and achieves SOTA performance \cite{sun2019deep}.

In this paper, we advance the research by studying the role of context information. Specifically, a novel method named progressive context refinement (PCR) is proposed for human keypoint detection. First, we devise a simple but effective context-aware module (CAM) that can efficiently integrate spatial and channel context information to aid feature learning for locating hard keypoints. Then, we construct the PCR model by stacking several CAMs sequentially with shortcuts and employ multi-task learning to progressively refine the context information and predictions. Besides, to maximize PCR's potential for the aforementioned hard case inference, we propose a hard-negative person detection mining strategy together with a joint-training strategy by exploiting the unlabeled coco dataset and external dataset.

The contributions of this work are as follows:

$\bullet$ We devise a simple but effective context-aware module (CAM) which serves as the key component for learning both spatial and channel context information.

$\bullet$ We propose a progressive context refinement model to predict and refine the keypoint locations gradually under the multi-task learning framework.

$\bullet$ We propose several efficient training strategies to guide PCR to deal with false-positive person detections and learn better feature representation from more samples.

$\bullet$ We set the new state-of-the-art result on the challenging COCO keypoint detection benchmark.

\section{Progressive Context Refinement for Human Keypoint Detection}
\label{sec:method}
In this paper, we tackle the multi-person pose estimation problem by following a top-down scheme. First, a human detector is used to detect the bounding box for each person instance. Then, PCR detects keypoints for each person instance. Finally, after aggregating the detections using Object Keypoint Similarity (OKS)-based Non-Maximum Suppression (NMS), we obtain the final pose estimation. The details of PCR are presented as follows.

\subsection{Context-Aware Module}
\label{subsec:cam}

\begin{figure}[t]
\centering
\includegraphics[width=1\linewidth]{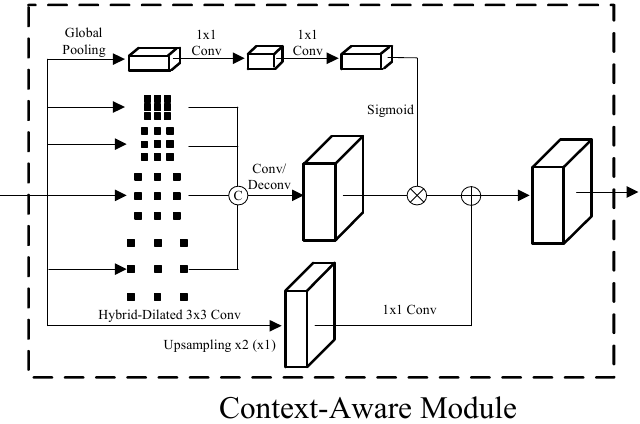}
\caption{The structure of Context-Aware Module (CAM).}
\label{fig:CAM}
\end{figure}

The CAM contains three branches: 1) a residual branch to retain the learned features from the previous stage; 2) a channel context extraction branch which inherits the idea of squeeze-and-excitation (SE) network \cite{hu2018squeeze} to calibrate the channel-wise features and capture the global contextual information; and 3) a hybrid-dilated convolutional(HDC) branch which inherits the idea of atrous spatial pyramid pooling (ASPP) \cite{chen2018deeplab} to capture multi-scale spatial contextual information within different receptive fields.

In the SE branch, the feature maps first go through a global pooling layer. Then, the obtained feature vector is fed into a bottle-neck layer with $1 \times 1$ convolutions. The feature dimension is reduced to $1 \times 1 \times {{{C_k}} \mathord{\left/
 {\vphantom {{{C_k}} 4}} \right.
 \kern-\nulldelimiterspace} 4}$, where $k$ is the index of CAM. Then, it is fed into a subsequent $1 \times 1$ convolutional layer to increase the feature dimension to $1 \times 1 \times C_{k}$, where $C_k$ represent the output feature channels. A sigmoid function is used to squeeze the feature vector $f_k^{SE}$ into the range $\left[ {0,1} \right]$, which is then used to calibrate the output feature maps from the HDC branch.

In the HDC branch, the feature maps go through four $3 \times 3$ convolutional layers with different dilated rates, \emph{i.e.}, 1, 2, 3, and 4. Each convolutional layer has ${{{C_k}} \mathord{\left/
 {\vphantom {{{C_k}} 4}} \right.
 \kern-\nulldelimiterspace} 4}$ kernels. These feature maps are then concatenated and fed into a deconvolutional layer of stride 2 or a convolutional layer of stride 1. The output feature maps $f_k^{HDC}$
are of size $H_k \times W_k \times C_k$, where $H_k$ and $W_k$ denote the height and width.

In the residual branch, feature maps from the previous stage are first up-sampled 2 times before being fed into a $1 \times 1$ convolutional layer to output feature maps $f_k^{RES}$ of size $H_k \times W_k \times C_k$. If the stride in the HDC branch is 1, there is no up-sampling.

Then, the output of the $k^{th}$ CAM can be calculated as:
\begin{equation}
f_k^{CAM} = f_k^{SE} \odot f_k^{HDC} + f_k^{RES},
\label{eq:feat_cam}
\end{equation}
where $\odot$ denotes the channel-wise multiplication. BN is used after each convolutional layer and deconvolutional layer. ReLU is used after the first convolutional layer in the SE branch, after all the convolutional layers in the HDC branch, and after the output of CAM.

\subsection{Progressive Context Refinement}
\label{subsec:pcr}

\begin{figure}[t]
\centering
\includegraphics[width=1\linewidth]{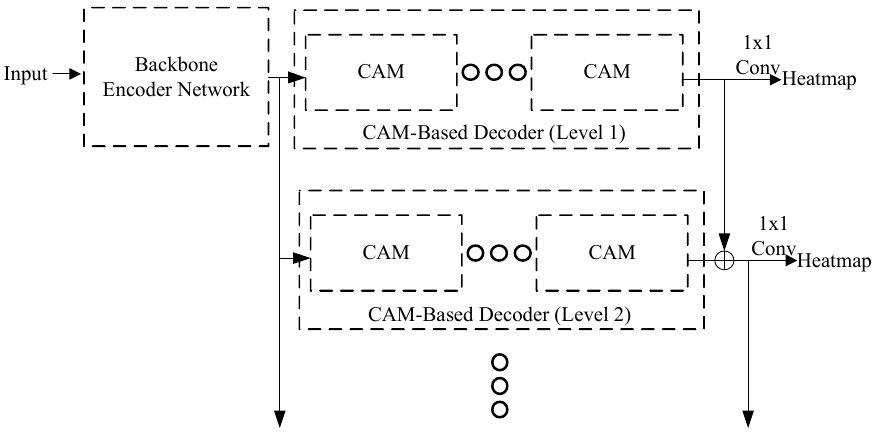}
\caption{The structure of the proposed Progressive Context Refinement (PCR) model.}
\label{fig:PCR}
\end{figure}

First, we construct a CAM-based decoder by stacking $K$ CAMs sequentially after the backbone encoder. Then, a $1\times1$ convolutional layer is employed as the prediction layer to output the heatmaps. This structure forms a single level encoder-decoder keypoint detection model, where context features are aggregated within each CAM and refined gradually in the following CAMs.

Then, we construct the PCR model by stacking $L$ CAM-based decoders in parallel, where they share the same encoded features as input. At each level, the output features are fused together with their counterparts from all previous levels by element-wise sum, which are then used for predicting the heatmaps. In this way, any subsequent level of decoder learns residual features compared to the previous levels and refines the heatmaps progressively. Mathematically, it can be formulated as:
\begin{equation}
{h_l} = {\varphi _l}\left( {\sum\limits_{i = 1,...l} {f_{iK}^{CAM}} } \right), l = 1,...,L
\label{eq:prediction}
\end{equation}
where $f_{iK}^{CAM}$ is the output features from the $K^{th}$ CAM at the $i^{th}$ level of decoder, ${\varphi _l}\left(  \cdot  \right)$
 denotes learned mapping function by the prediction layer at the $l^{th}$ level, $h_l$ denotes the predicted heatmaps.

\subsection{Efficient Training Strategies}
\label{subsec:trainStrategies}

\textbf{Hard-Negative Person Detection Mining (HNDM)}: Top-down approaches detect person instances before detecting keypoints on them. Although modern detection models have achieved a good detection performance, they may still produce some false positive detections due to occlusion, similar appearances, etc. To address this issue, we propose a hard-negative person detection mining strategy. After obtaining the person detections, we filter out those detections with high scores but no intersections with ground truth person instances, $i.e.$, hard-negative detections for the subsequent keypoint detection phase. Their heatmaps are set to zero. Using these training samples drives PCR to predict no keypoints on those false ``person instances''.

\textbf{Joint Training on Unlabeled COCO dataset and External Dataset}: To increase pose diversity in the training samples, we leverage the MS COCO unlabeled dataset and the external dataset named AI Challenger (AIC) \cite{AIC2017}. We train a keypoint detector using a ResNet-152 backbone. For the detected person instances on the unlabelled dataset, we keep all keypoints with scores above 0.9 as the pseudo-annotations and treated the rest as unlabeled. Only 14 categories are annotated in the AIC dataset. Instead of using transfer learning, we keep their common annotations with the same categories as the COCO dataset and discarding the others. We add the samples from the unlabelled COCO dataset and AIC dataset as described above into the original COCO training set and use them to train PCR model jointly.

\section{Experiments}
\label{sec:experiment}

\subsection{Experimental settings}
\label{subsec:settings}
\textbf{Datasets}: The COCO Keypoint detection dataset is split into training, minival, test-dev, and test-challenge sets \cite{lin2014microsoft}. The training set includes 118k images and 150k person instances, the minival dataset includes 5000 images, and the test set includes 40k images, 20k each in test-dev and test-challenge. It also provides an unlabeled dataset containing 123k images. 110k person instances and corresponding keypoints are detected using the method described in Section~\ref{subsec:trainStrategies}. The external dataset from AIC \cite{AIC2017} contains a training set with 237k images and 440k person instances and a validation set with 3000 images. We report the main results according to mean average precision (AP) over 10 object keypoint similarity (OKS) thresholds \cite{lin2014microsoft}.

\textbf{Implementation details}: The feature dimension $C_i$ of each CAM was set to 256 for ResNet-50 backbone, 128, 96, 64 for ResNet-152 backbone, 48 for HRNet backbone. All other hyper-parameters were set by following \cite{xiao2018simple, sun2019deep}. We used the detection results on the minival and test-dev sets released in \cite{xiao2018simple} if not specified.

\subsection{Main Results}
\label{subsec:mainResults}

\begin{table}[htbp]
\footnotesize
  \centering
  \caption{Comparisons of PCR and SOTA methods on the COCO test-dev set. *: external data, +: ensemble model.}
    \begin{tabular}{p{3.05cm}p{0.35cm}<{\centering}p{0.5cm}<{\centering}p{0.6cm}<{\centering}p{0.35cm}<{\centering}p{0.35cm}<{\centering}p{0.35cm}<{\centering}}
    \toprule
      Method    & $AP$    & $AP^{@.5}$  & $AP^{@.75}$  & $AP^M$   & $AP^L$   & $AR$ \\
    \midrule
    Baseline \cite{xiao2018simple}(R152) &  73.7     &  91.9     &  81.1     &  70.3     &  80.0     & 79.0 \\
    Baseline+* \cite{xiao2018simple}(R152) &  76.5     &  92.4     &  84.0     &  73.0     & 82.7     & 81.5 \\
    HRNet-W48 \cite{sun2019deep} & 75.5     &  92.5     &  83.3     &  71.9     &  81.5    & 80.5 \\
    HRNet-W48* \cite{sun2019deep} & 77.0    &  92.7     &  84.5     &  73.4     &  83.1    & 82.0 \\
    Megvii+* \cite{li2019rethinking}\footnotemark[1] &78.1  &\textbf{94.1}       &85.9       &74.5       &83.3       &83.1 \\
    \hline
    \textbf{PCR (R152)} &  75.6    &  92.6     &  83.3     &  71.8     &  81.1    & 80.7 \\
    \textbf{PCR* (R152)} &  77.1    &  93.0     &  84.7     &  73.2     &  82.7    & 82.1 \\
    \textbf{PCR* (HRNet-48)} &  77.9    &  93.5     &  85.0     &  73.9     &  83.5    & 82.9 \\
    \textbf{PCR+* (R152,HRNet-48)} &  \textbf{78.9}    &  93.8     &  \textbf{86.0}     &  \textbf{75.0}     &  \textbf{84.5}    & \textbf{83.6} \\
    \bottomrule
    \end{tabular}%
  \label{tab:SOTA_testDev}%
\end{table}%

\footnotetext[1]{The champion of the 2018 COCO Keypoint Challenge.}

The results of PCR and SOTA methods on the COCO test-dev set are summarized in Table~\ref{tab:SOTA_testDev}. The input size is $384\times288$. PCR outperforms its baseline model by a healthy margin, $i.e.$, 1.9 for Baseline \cite{xiao2018simple} and 0.9 for HRNet-48. It is noteworthy that our single model \emph{PCR* (HRNet-48)} achieves comparable performance with the winner of the 2018 COCO Keypoint Challenge \emph{Megvii+*}. Our final ensemble model \emph{PCR+* (R152,HRNet-48)} sets a new SOTA on this benchmark, $i.e.$, 78.9.

For the final submission to the 2019 keypoint detection challenge, we use a detector with a person AP of 60.6 on the test-dev set. The results are listed in Table~\ref{tab:resultsTestCha}.

\begin{table}[htbp]
\footnotesize
  \centering
  \caption{The final result of PCR on the COCO challenge.}
    \begin{tabular}{p{3.05cm}p{0.35cm}<{\centering}p{0.5cm}<{\centering}p{0.6cm}<{\centering}p{0.35cm}<{\centering}p{0.35cm}<{\centering}p{0.35cm}<{\centering}}
    \toprule
      Method    & $AP$    & $AP^{@.5}$  & $AP^{@.75}$  & $AP^M$   & $AP^L$   & $AR$ \\
    \midrule
    \textbf{PCR+* (test-dev)}\footnotemark[2] & 78.9    &  93.8     &  86.0     &  74.9     &  84.5    & 83.4 \\
    \textbf{PCR+* (test-challenge)} &  75.5    &  92.3     &  82.1     &  69.9     &  82.8    & 81.1 \\
    \bottomrule
    \end{tabular}%
  \label{tab:resultsTestCha}%
\end{table}%

\footnotetext[2]{The scores are slightly different from Table~\ref{tab:SOTA_testDev} due to using different person detection results and the limit of uploaded file size.}

\subsection{Ablation Study}
\label{subsec:ablation}

\begin{table}[htbp]
\footnotesize
  \centering
  \caption{Ablation study on the components of PCR. AD: auxiliary task after the penultimate CAM. AP/AR: mean average precision/recall on COCO minival set.}
    \begin{tabular}{p{2.6cm}p{0.4cm}<{\centering}p{0.5cm}<{\centering}p{0.4cm}<{\centering}p{0.4cm}<{\centering}p{0.4cm}<{\centering}}
    \toprule
   Method & SE   & HDC    & AD    & \emph{AP}    & \emph{AR} \\
    \midrule
   Baseline \cite{xiao2018simple} &  &  &   & 70.4  & 76.3 \\
   HRNet-W32 \cite{sun2019deep} &  &  &   &  74.4    & 79.8 \\
    \hline
   PCR(R50) &\checkmark  &  &   & 72.8  & 78.7 \\
   PCR(R50) &  & \checkmark      &       & 72.6  & 78.5 \\
   PCR(R50) &    &    &\checkmark       & 73.0    & 78.7 \\
   PCR(R50) & \checkmark    & \checkmark & \checkmark     & 73.8  & 79.3 \\
   \textbf{PCR(HRNet-W32)}  & \checkmark &\checkmark   &\checkmark  &  \textbf{75.8}       & \textbf{81.1}  \\
    \bottomrule
    \end{tabular}%
  \label{tab:ablation_component}%
\end{table}%

\begin{table}[htbp]
\footnotesize
  \centering
  \caption{Comparisons of PCR trained with the different strategies described in Section ~\ref{subsec:trainStrategies}. A: AIC, C: COCO, H: HNDM, U: Unlabelled COCO.}
    \begin{tabular}{p{1.6cm}<{\centering}p{0.75cm}<{\centering}p{0.75cm}<{\centering}p{1.5cm}<{\centering}p{1.5cm}<{\centering}}
    \toprule
    PCR(R50) & A$\to$C   & AC$\to$C  & ACH$\to$CH & ACHU$\to$CH \\
     \midrule
    \emph{AP} & 74.8  & 75.0    & 75.3  & 75.6 \\
    \emph{AR} & 80.1  & 80.4  & 80.4  & 80.7 \\
    \bottomrule
    \end{tabular}%
  \label{tab:ablation_trainingStrategy}%
\end{table}%

The result of the ablation study on the components of PCR are listed in Table~\ref{tab:ablation_component}. The backbone network is ResNet-50 (R50, K=3, L=1) and HRNet-32 (K=1, L=1), and the input size was $256 \times 192$. As can be seen, each component achieved gains over the baseline model \cite{xiao2018simple}. Besides, the complementarity between these components in CAM leads to better results as validated in the last two rows.

The results of using different training strategies described in Section ~\ref{subsec:trainStrategies} are summarized in Table~\ref{tab:ablation_trainingStrategy}, where A$\to$C denotes the transfer learning strategy, AC$\to$C denotes the joint-training strategy training PCR both AIC and COCO datasets then fine-tuning it on COCO dataset. Other symbols have a similar meaning. As can be seen, our HNDM and joint-training strategies on AIC and unlabelled COCO datasets consistently improve the performance.

\section{Conclusion}
\label{sec:conclusion}
In this paper, we propose a novel progressive context refinement model (PCR) for human keypoint detection. It builds on the simple but effective context-aware module (CAM) which efficiently integrates spatial and channel context information gradually. Under the multi-task learning framework, PCR stacks several CAMs sequentially to refine the context information within a single decoder and stacks several such decoders in parallel to fuse the learned features and refine the predictions progressively. Effective training strategies including hard-negative person detection mining and joint-training on the unlabeled coco dataset and external dataset are proposed. Experiments demonstrate that PCR outperforms representative state-of-the-art (SOTA) methods and sets a new SOTA on MS COCO benchmark.

{\small
\bibliographystyle{ieee_fullname}
\bibliography{2019ICCV_COCO_PCR}

\begin{thebibliography}{10}\itemsep=-1pt

\bibitem{AIC2017}
Ai challenger human keypoint detection dataset.
\newblock \url{https://challenger.ai/competition/keypoint/}.

\bibitem{chen2018deeplab}
Liang-Chieh Chen, George Papandreou, Iasonas Kokkinos, Kevin Murphy, and Alan~L
  Yuille.
\newblock Deeplab: Semantic image segmentation with deep convolutional nets,
  atrous convolution, and fully connected crfs.
\newblock {\em IEEE transactions on pattern analysis and machine intelligence},
  40(4):834--848, 2018.

\bibitem{chen2018cascaded}
Yilun Chen, Zhicheng Wang, Yuxiang Peng, Zhiqiang Zhang, Gang Yu, and Jian Sun.
\newblock Cascaded pyramid network for multi-person pose estimation.
\newblock In {\em CVPR}, pages 7103--7112, 2018.

\bibitem{hu2018squeeze}
Jie Hu, Li Shen, and Gang Sun.
\newblock Squeeze-and-excitation networks.
\newblock In {\em Proceedings of the IEEE conference on computer vision and
  pattern recognition}, pages 7132--7141, 2018.

\bibitem{li2019rethinking}
Wenbo Li, Zhicheng Wang, Binyi Yin, Qixiang Peng, Yuming Du, Tianzi Xiao, Gang
  Yu, Hongtao Lu, Yichen Wei, and Jian Sun.
\newblock Rethinking on multi-stage networks for human pose estimation.
\newblock {\em arXiv preprint arXiv:1901.00148}, 2019.

\bibitem{lin2014microsoft}
Tsung-Yi Lin, Michael Maire, Serge Belongie, James Hays, Pietro Perona, Deva
  Ramanan, Piotr Doll{\'a}r, and C~Lawrence Zitnick.
\newblock Microsoft coco: Common objects in context.
\newblock In {\em ECCV}, pages 740--755. Springer, 2014.

\bibitem{newell2016stacked}
Alejandro Newell, Kaiyu Yang, and Jia Deng.
\newblock Stacked hourglass networks for human pose estimation.
\newblock In {\em ECCV}, pages 483--499. Springer, 2016.

\bibitem{sun2019deep}
Ke Sun, Bin Xiao, Dong Liu, and Jingdong Wang.
\newblock Deep high-resolution representation learning for human pose
  estimation.
\newblock {\em arXiv preprint arXiv:1902.09212}, 2019.

\bibitem{toshev2014deeppose}
Alexander Toshev and Christian Szegedy.
\newblock Deeppose: Human pose estimation via deep neural networks.
\newblock In {\em CVPR}, pages 1653--1660, 2014.

\bibitem{xiao2018simple}
Bin Xiao, Haiping Wu, and Yichen Wei.
\newblock Simple baselines for human pose estimation and tracking.
\newblock In {\em ECCV}, pages 466--481, 2018.

\end{thebibliography}

}

\end{document}